\documentclass{article}

\usepackage{arxiv}
\usepackage{graphicx}

\usepackage[utf8]{inputenc} % allow utf-8 input
\usepackage[T1]{fontenc}    % use 8-bit T1 fonts
\usepackage{hyperref}       % hyperlinks
\usepackage{url}            % simple URL typesetting
\usepackage{booktabs}       % professional-quality tables
\usepackage{amsfonts}       % blackboard math symbols
\usepackage{nicefrac}       % compact symbols for 1/2, etc.
\usepackage{microtype}      % microtypography
\usepackage{graphicx}
\usepackage{amsmath}
\usepackage{amssymb}

\usepackage{algorithm}
\usepackage{algpseudocode}
\usepackage{comment}

\DeclareMathOperator*{\argmax}{argmax}
\DeclareMathOperator*{\argmin}{argmin}

\title{Approximating Solutions to the Knapsack Problem using the Lagrangian Dual Framework}

\author{Mitchell Keegan\\
	University of Queensland \\
	\texttt{m.keegan@uqconnect.edu.au} \\
	\And
	Mahdi Abolghasemi\\
	University of Queensland \\
	\texttt{m.abolghasemi@uq.edu.au} \\
}

\date{}

\renewcommand{\headeright}{}
\renewcommand{\undertitle}{The Version of Record of this contribution is published in Lecture Notes in Computer Science Volume 14471, and is available online at https://doi.org/10.1007/978-981-99-8388-9\_37}
\renewcommand{\shorttitle}{}

\begin{document}
\maketitle

\begin{abstract}
The Knapsack Problem is a classic problem in combinatorial optimisation. Solving these problems may be computationally expensive. Recent years have seen a growing interest in the use of deep learning methods to approximate the solutions to such problems. A core problem is how to enforce or encourage constraint satisfaction in predicted solutions. A promising approach for predicting solutions to constrained optimisation problems is the Lagrangian Dual Framework which builds on the method of Lagrangian Relaxation. In this paper we develop neural network models to approximate Knapsack Problem solutions using the Lagrangian Dual Framework while improving constraint satisfaction. We explore the problems of output interpretation and model selection within this context. Experimental results show strong constraint satisfaction with a minor reduction of optimality as compared to a baseline neural network which does not explicitly model the constraints. 
\end{abstract}
\section{Introduction}

The Knapsack Problem (KP) is a classic problem in combinatorial optimisation (CO). The core statement is as follows; there are $n$ items, each with weight $w_i$ and value $v_i$. The goal is to select a set of items which maximise the total value, such that the total weights of the chosen items is not greater than some capacity $W$. It is formally defined as an integer programming (IP) problem as shown below:

\begin{equation}
    \begin{aligned}
        x^* = \argmax_x & \sum_{i=1}^n x_iv_i \\
        \text{s.t.} \:\: &  \sum_{i=1}^n x_iw_i \leq W \\   
        & x_i \in \{0,1\}, \: i=1,...,n
    \end{aligned}
\end{equation}

where $x_i=1$ if the $i^{th}$ item is chosen or $x_i=0$ if not. KP and its variants find applications when limited resources must be allocated efficiently. Examples include cutting stock problems, investment allocation, and transport and logistics. There exists effective exact and approximate algorithms to solve KP including dynamic programming, branch and bound, and a variety of metaheuristic algorithms. A detailed treatment of these algorithms for KP and its many variants can be found in \cite{pisinger04a}. A common drawback of these methods for KP and other CO problems is that the solving time may be too slow in applications where solutions are required under strict time constraints. A promising remedy is to use fast inference machine learning models to approximate solutions \cite{abolghasemi2021effectively,abolghasemi2023machine,abolghasemi2021state}.

The idea of applying machine learning to CO is not a new one, although recent years have seen a resurgence in interest with the rise of deep learning. An early example in 1985 was the use of Hopfield Networks to approximate the objective of the Travelling Salesman Problem \cite{hopfield85}. The last decade has seen explosive advances in machine learning, with progress in deep learning providing practical solutions to problems in the field of computer vision and natural language processing (among many others) which were once considered incredibly difficult or intractable. This has naturally led to renewed interest in the applications of machine learning to CO. 

This paper focuses on using supervised deep learning to predict solutions to KP. For broad reviews of recent advancements in the applications of machine learning to CO we refer to \cite{bengio18}, \cite{abolghasemi22}, and \cite{kotary21}. A core challenge in machine learning for CO is data generation. Solving CO problems is often computationally expensive which makes generating target labels for large datasets difficult. We sidestep these challenges in this paper by focusing on KP for which we can generate solutions with reasonable speed, but this problem represents an interesting area of research \cite{kotary21b}. For the purpose of approximating solutions to constrained optimisation problems, a key weakness of neural networks is the inability to enforce constraint satisfaction on the model outputs.

Several models have been developed to address this weakness. One approach proposes using an iterative training routine in which a learner step that trains a model as usual is interleaved with a master step that adjusts the target labels to something closer to the predicted solution while remaining feasible \cite{detassis21}. It makes no assumptions about the nature of the constraints and permits any supervised ML model in the learner step. Another approach is Deep Constraint Completion and Correction (DC3) \cite{donti21}. DC3 uses differentiable processes, named completion and correction, to enforce the feasibility of solutions during training. It showed strong results with a high degree of constraint satisfaction on the AC optimal power flow problem (AC-OPF), which is a continuous optimisation problem. A third approach to integrating constraints is based on Lagrangian Relaxation.

Lagrangian Relaxation is a method of solving constrained optimisation problems by relaxing constraints into the objective function scaled by Lagrange Multipliers. \cite{fontaine14} presents a generalisation of Lagrangian Relaxation and related concepts which allows it to be applied to arbitrary optimisation models. The Lagrangian Dual Framework \cite{fioretto20ACOPF,fioretto20LDFC} builds on this approach, using Lagrangian relaxation to encourage constraint satisfaction in neural network models during training. The LDF has been applied with encouraging results to the problems of AC-OPF \cite{fioretto20ACOPF}, constrained predictors where constraints between samples are present \cite{fioretto20LDFC}, and the job shop scheduling problem \cite{kotary22}.

In this paper, we apply the LDF to the Knapsack Problem. This is not the first attempt at predicting KP solutions using neural networks. For example GRU, CNN and feedforward neural networks have been designed for this purpose \cite{nomer20}, but these models did not incorporate constraint satisfaction into the training procedure. Some research has been done to derive theoretical bounds on the depth and width of an RNN designed to mimic a dynamic programming algorithm for both exact and approximate solutions \cite{hertrich21}. Directly incorporating constraints in the training process should theoretically improve performance since it explicitly models the constraints within the learning framework, instead of only appearing implicitly in the training data.

In this paper we develop three neural network models to approximate solutions to KP, one is a baseline fully connected network which is compared to another fully connected network that models the constraints using the LDF. A third model uses the LDF but also uses the baseline neural network as a pre-trained model. We explore implementation details for applying the LDF to KP (and by extension IP problems), experimentally investigate the trade-off between the optimality of approximated solutions against constraint satisfaction, and discuss how to approach hyperparameter tuning and model selection in the context of this trade-off.

\section{Lagrangian Dual Framework}
We implement the LDF initially proposed in \cite{fioretto20LDFC} to learn a model which approximates solutions to the Knapsack problem. This section restates the formulation of the LDF as relevant to KP. Consider a general constrained optimisation problem with inequality constraints:
\begin{equation}
    \begin{aligned}
        \mathcal{O} = & \argmin_y & & f(y) \\
        & \text{subject to} & & g_i(y) \leq 0, \; i=1, \ldots ,m \\
    \end{aligned}
\end{equation}

The violation-based Lagrangian function is:

\begin{equation}
    f_\lambda(y) = f(y) + \sum_{i=1}^{m} \lambda_i \text{max}(0,g_i(y))
\end{equation}

where $\lambda_i \geq 0$ denote the Lagrange multipliers associated with the inequality constraints. Another approach is to consider the satisfiability-based Lagrangian function:

\begin{equation}
    f_\lambda(y) = f(y) + \sum_{i=1}^{m} \lambda_i g_i(y)
\end{equation}

Both of these can be generalised by considering the function $\nu(g_i)$ which returns either the constraint satisfiability or the degree of constraint violation as required. 

\begin{equation}
    f_\lambda(y) = f(y) + \sum_{i=1}^{m} \lambda_i \nu(g_i(y))
\end{equation}

In this paper $\nu(g_i)$ will always refer to the constraint violation degree $\text{max}(0,g_i(y))$. The Lagrangian Relaxation is then:

\begin{equation}
    LR_\lambda = \argmin_y f_\lambda (y)
\end{equation}

for some set of Lagrangian multipliers $\lambda = \{\lambda_1 , \hdots , \lambda_m\}$. The solution forms a lower bound on the original constrained problem, i.e. $f(LR_\lambda) \leq f(\mathcal{O})$.

To find the strongest Lagrangian relaxation of $\mathcal{O}$, we can find the best set of Lagrangian multipliers using the Lagrangian dual.

\begin{equation}
    LD = \argmax_{\lambda \geq 0} f(LR_\lambda)
\end{equation}

The LDF leverages the Lagrangian Relaxation to improve constraint satisfaction in neural network models \cite{fioretto20ACOPF}. Consider parametrising the original optimisation problem by the parameters of the objective function and constraints:

\begin{equation}
    \begin{aligned}
        \mathcal{O}(d) = & \argmin_y & & f(y,d) \\
        & \text{subject to} & &  g_i(y,d) \leq 0, \; i=1, \ldots ,m 
    \end{aligned}
\end{equation}

The goal is the learn some parametric model $\mathcal{M}_w$ with weights $w$ such that $\mathcal{M}_w \approx \mathcal{O}$. In this paper $\mathcal{M}_w$ is always a feedforward neural network.

A set of data is denoted $D = \{(d_l,y_l=\mathcal{O}(d_l)\}$ for $l=1,\ldots,n$. The learning problem is to solve:

\begin{equation}
    \begin{aligned}
        w^*= & \argmin_w & & \sum_{l=1}^n\mathcal{L}(\mathcal{M}_w(d_l),y_l) \\
        & \text{subject to} & &  g_i(\mathcal{M}_w(d_l),d_l) \leq 0, \; i=1, \ldots ,m,  \; l=1, \ldots ,n  
    \end{aligned}
\end{equation}

for some loss function $\mathcal{L}$. In essence this states that $\mathcal{M}_w$ should have weights such that it minimises the loss over all samples, while being such that the output satisfies the constraints for all samples. To this end, relax the constraints into the loss function to form the Lagrangian loss function:

\begin{equation}
    \mathcal{L}_\lambda (\hat{y}_l ,y_l,d_l) = \mathcal{L}(\hat{y}_l,y_l) + \sum_{i=1}^m \lambda_i \nu (g_i(\hat{y}_l,d_l))
\end{equation}

where $\hat{y}_l = \mathcal{M}_w(d_l)$. For a given set of Lagrange multipliers $\lambda$, the Lagrangian relaxation is:

\begin{equation}
        w^*(\lambda)= \argmin_w \sum_{l=1}^n\mathcal{L}_\lambda (\mathcal{M}_w(d_l),y_l,d_l) \\
\end{equation}

The solution is an approximation $\mathcal{M}_{w^*(\lambda)}$ of $\mathcal{O}$. To find a stronger Lagrangian relaxation, the Lagrangian Dual is used to compute the optimal multipliers:

\begin{equation}
        \lambda^*= \argmax_\lambda \min_w \sum_{l=1}^n\mathcal{L}_\lambda (\mathcal{M}_w(d_l),y_l,d_l) \\
\end{equation}

The LDF implements subgradient optimisation to iteratively solve for $w$ and $\lambda$. The training process is summarised in Algorithm \ref{alg:LDF}. It takes in the following inputs: Training data $D$, number of training epochs $n_{epochs}$, Lagrangian step size $s_i$ for $i=1,...,m$,  and initial Lagrange multipliers $\lambda_i^0$ for $i=1,...,m$.

\begin{algorithm}
\caption{LDF Algorithm}\label{alg:LDF}
\;\;\;\:\textbf{Input:} $D$, $n_{epochs}$, $s_i$, $\lambda_i^0$
\begin{algorithmic}
\For{$k=0,1,...,n_{epochs}$}
    \ForAll{$(y_l,d_l) \in D$}
        \State $\hat{y} \gets \mathcal{M}_w(d_l)$
        \State $w \gets w - \alpha \nabla_w \mathcal{L}_{\lambda^k} (\hat{y},y_l,d_l)$
    \EndFor
    \State $\lambda_i^{k+1} \gets \lambda_i^{k} + s_i \sum_{l=1}^n \nu_i(g_i(\hat{y},d_l)), \;\;\; i=1,...,m$
\EndFor
\end{algorithmic}
\end{algorithm}

\begin{comment}
    \begin{algorithm}
\caption{LDF Algorithm}\label{alg:LDF}
\;\;\;\:\textbf{Input:} $D = \{(d_l,y_l=\mathcal{O}(d_l)\}$ for $l=1,\ldots,n$, $s_k$
\begin{algorithmic}
\For{$k=0,1,...,n_{epochs}$}
    \ForAll{$(y_l,d_l) \in D$}
        \State $\hat{y} \gets \mathcal{M}_w(d_l)$
        \State $w \gets w - \alpha \nabla_w \mathcal{L}_{\lambda^k} (\hat{y},y_l,d_l)$
    \EndFor
    \State $\lambda_i^{k+1} \gets \lambda_i^{k} + s_k \sum_{l=1}^n \nu_i(g_i(\hat{y},d_l)), \;\;\; i=1,...,m$
\EndFor
\end{algorithmic}
\end{algorithm}
\end{comment}

\section{Experiment setup}
In this section, we describe the approach taken to approximate the solutions to the Knapsack Problem using neural networks within the Lagrangian Dual Framework. We  provide the details of our methods for instance generation, introduce our approach for decoding and evaluating solutions predicted by our models, and discuss the network architectures and training procedures.

\subsection{Instance Generation}
Instances should be generated for training and testing which represent a wide variety of adequately difficult cases. This paper uses the size of the knapsack capacity relative to the total item weights as a proxy for instance difficulty. We generated $30,000$ instances each with $n=500$ items, from which we used $24,000$ instances for training while $3000$ were set aside for testing and validation sets. The item weights and values are uncorrelated and uniformly distributed on $[0,1]$. A method proposed in \cite{pisinger02} is used to generate the instance capacities, where the capacity of the $j^{th}$ instance is set to $W_j = \frac{j}{S+1}\sum_{i=1}^nw_i$, $S$ being the total number of instances. This produces data with a full coverage of instance capacities and difficulties, from instances in which the optimal solution have few items to those which have almost all items. Target labels were generated for each instance using Gurobi 9.5.2 on a PC running Ubuntu Linux with an Intel i5-11400 processor.

\subsection{Output Decoding and Evaluation}
The output of the neural network is a vector in $\mathbb{R}^n$, where the $i^{th}$ element is a logit corresponding to the $i^{th}$ weight. Binary cross entropy is used for the label portion of the loss function $\mathcal{L}(\hat{y}_l,y_l$). 

Care must be taken in using the model outputs to evaluate the knapsack constraint. At training time the logit outputs must be mapped to binary decision variables. A naive approach is to apply a sigmoid at the output to scale it into the range $[0,1]$ and then round to zero or one. The problem is that the derivative of the round function is zero everywhere, meaning that informative gradients will not flow back from the constraint evaluation. Instead, a surrogate gradient is substituted in during the backwards pass, similar to methods used to deal with uninformative gradients in spiking neural networks such as in \cite{eshraghian21}. The gradient of the sigmoid function centred at $0.5$ is used as a surrogate with the form: 

\begin{equation}
    \frac{d\sigma}{dx} = \frac{ke^{-k(x-0.5)}}{(e^{-k(x-0.5)} + 1)^2}
\end{equation}

\begin{figure}[h]
\centering
\includegraphics[width=8cm]{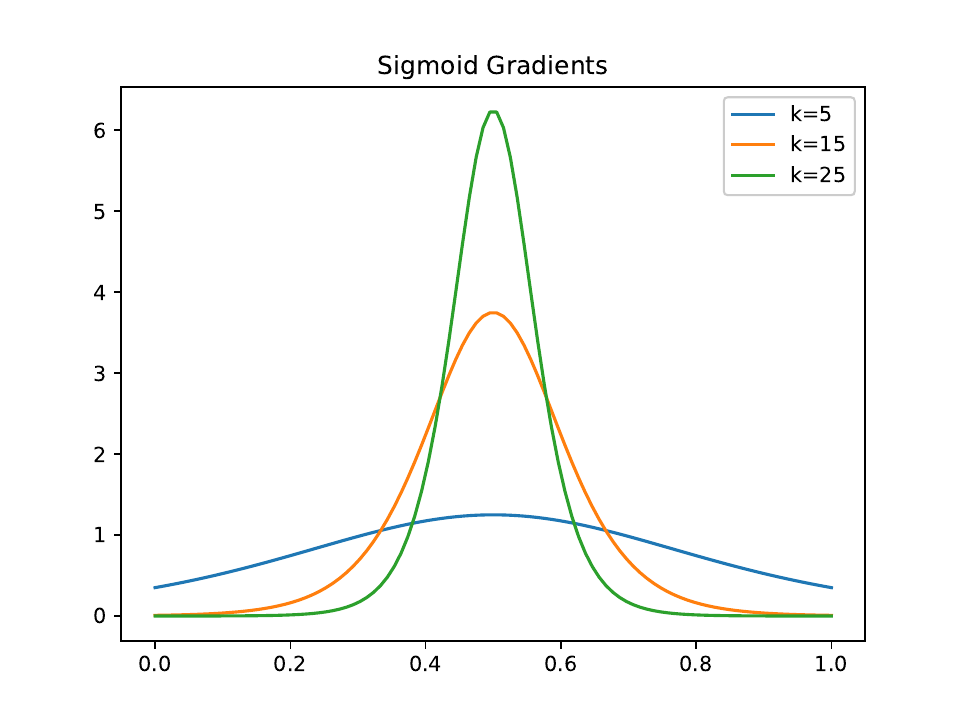}
\caption{Surrogate Gradient} \label{fig KP: Surrogate Gradient}
\end{figure}

The parameter $k$ affects how tightly the function is distributed around $0.5$ as shown in Figure \ref{fig KP: Surrogate Gradient}. The gradient will be higher when outputs are near $0.5$ and tend towards zero near $0$ and $1$. This reflects the fact that near $0.5$ small changes in the input can cause the output of the round function to jump between $0$ and $1$.

Rounding is also used at inference time to decode the model outputs. Another decoding scheme as proposed in \cite{nomer20}, is a greedy algorithm which ensures the capacity constraint is satisfied. For general IP problems, it may not be obvious how to correct infeasible solutions, or it may be computationally expensive to do so. For this reason, although reasonable decoding schemes exist, this paper chooses to focus on a minimal decoding which simply rounds the outputs to investigate performance under minimal output post-processing. As in \cite{nomer20} the approximation ratio (AR) is used as a metric to evaluate the model. For a set of instances of size $S$ it is defined as: 

\begin{equation*}
    AR = \frac{1}{S} \sum_{j=1}^S max \left(\frac{f^*(x_j)}{f(x_j)},\frac{f(x_j)}{f^*(x_j)} \right)
\end{equation*}

where $f^*(x_j)$ and $f(x_j)$ are the predicted and true optimal objective values for the $j^{th}$ instance. The approximation ratio is not well defined for instances where either $f^*(x_j)$ or $f(x_j)$ is equal to zero, but not both. Here we define the approximation ratio to be equal to two whenever this occurs. While arbitrary, the idea is that a strong predictor should have an approximation ratio approaching one and this represents a small penalty for incorrect predictions. Regardless, instances where one of either $f^*(x_j)=0$ or $f(x_j)=0$ are rare and should not significantly influence results.

While the approximation ratio is important, for the purpose of model validation it only evaluates the optimality of predictions without considering constraint satisfaction. Since the Lagrangian multipliers update on each epoch, the loss $\mathcal{L}_\lambda (\hat{y}_l ,y_l,d_l)$ may increase between epochs even as overall model performance is improving which makes it a flawed metric for comparing performance across epochs. To this end, we introduce the $\mu$-loss, which replaces the Lagrange multiplier in the loss function with fixed $\mu$ values as shown below:

\begin{equation}
    \mathcal{L}_\mu (\hat{y}_l ,y_l,d_l) = \mathcal{L}(\hat{y}_l,y_l) + \mu \nu (g_i(\hat{y}_l,d_l))
\end{equation}

In choosing a model which minimises the $\mu$-loss it should be possible to select for models which place higher relative priority on prediction optimality or constraint satisfaction as needed by varying $\mu$. In practice model selection will depend on the specific application and the relative importance of optimality and constraint satisfaction.

\subsection{Network Architecture and Training Procedure}
We have trained three neural network models for approximating solutions to the Knapsack Problem: the first was a fully connected neural network (FC) used as a baseline model, the second model used the LDF to model the constraints, the third model used the LDF but also used the FC baseline as a pre-trained model. Equivalently the pre-trained LDF could be viewed as fixing the Lagrange multipliers at zero for some fixed amount of epochs or until the model converges before allowing them to be updated. To our knowledge there are currently no other suitable methods in the literature to use as a baseline model. All models were trained using PyTorch v1.12.1 with Numpy v1.23.5 on a PC running Ubuntu Linux with an Intel i5-11400 processor\footnote{Code available at https://github.com/mitchellkeegan/knapsack-ldf}. 

An input to the network is the parameters for a single instance. For a KP instance with $n$ items, the parameters are the weights $w_i$, values $v_i$ and capacity W. We then construct the input as a vector $\left[ w_1,\hdots,w_n,v_1,\hdots,v_n,W \right] \in \mathbb{R}^{2n+1}$

All models had two hidden layers with widths 2048 and 1024, ReLU activation functions, and were trained using the Adam optimiser with default $\beta$ values. While training all models batch normalization was applied on both hidden layers, $k=25$ was set for the surrogate gradient, a batch size of 256 was used, and the Lagrange multiplier was initialised to $1$. In practice varying $k$ in the range $[5,35]$ did not have any noticeable effect. Training was performed for 500 epochs. We performed hyperparameter optimisation over the learning rate, maximum gradient norm, and Lagrangian step size. Values for hyperparameter optimisation are recorded in Table \ref{tab KP: Hyperparameters}. 

\begin{table}[h]
    \centering
    \begin{tabular}{|c|c|c|c|}
        \hline
        Model & Learning Rate & Lagrangian Step & Grad Norm \\
        \hline
        FC & \{$10^{-4},10^{-3}$\} & N/A & \{1,10\} \\ 
        LDF & \{$10^{-5}, 10^{-4},10^{-3}$\} & \{$10^{-8}, 10^{-7},10^{-6}, 10^{-5}, 10^{-4},10^{-3}$\} & \{0.5,1,10\} \\
        Pre-trained LDF & \{$10^{-4},10^{-3}$\} & \{$10^{-7},10^{-6}, 10^{-5}, 10^{-4},10^{-3}$\} & \{1,10\} \\
        \hline
    \end{tabular}
    \caption{Hyperparameter Optimisation Sets}
    \label{tab KP: Hyperparameters}
\end{table}

We selected models based on performance on the validation set. The chosen FC model was the one which minimised the AR, both LDF models were chosen to minimise the $1$-loss. All results are reported on the test set. The chosen FC model was trained with a learning rate of $10^{-3}$ and maximum gradient norm of $10$. The chosen LDF model was trained with a learning rate of $10^{-4}$, Lagrangian step size of $10^{-7}$, and maximum gradient norm of $0.5$. The chosen pre-trained LDF model was trained with a learning rate of $10^{-4}$, Lagrangian step size of $10^{-4}$, and maximum gradient norm of $10$.

A common problem in training the LDF model was exploding gradients originating from the constraint evaluation. The default weight initialisation produces an output with approximately half of the knapsack weights active. As a result, predictions early in the training process can significantly violate the capacity constraint in low capacity instances. Gradient clipping alleviates this, but it's unclear if this may distort the training process and reduce the relative importance of constraint satisfaction in learned models. This issue is much less pronounced in the pre-trained model but is still present. This made the pre-trained LDF models easier to train and more robust to changes in hyperparameters since they did not have the same degree of instability as the LDF models.

\section{Empirical Results}

Table \ref{tab KP: Baseline NN} reports results for the baseline FC model. All figures are reported as a percentage. For constraint violation, we report the percentage of instances in which the constraint was violated, and the average violation percentage. The average violation is calculated only over the instances in which the constraint is violated. It also filters out extreme outliers that can occur in very low capacity (those in which the knapsack capacity is extremely small compared to the sum of the item weights) instances by ignoring any instances in which the violation is more than six standard deviations from the mean violation. For the objective statistics, we report the percentage of instances in which the predicted objective was under and above the optimal objective, and the average undershoot and overshoot calculated only on the instances in which the predicted objective is under/over the optimal objective (labelled Avg-O and Avg-U). Lastly, the approximation ratio is reported. Performance is broken down into quintiles by instance capacity relative to total item weight denoted $\alpha = \frac{W_j}{\sum_{i=1}^n w_{ij}}$ where $w_{ij}$ and $W_j$ are the weight of the $i^{th}$ item and the knapsack capacity respectively in the $j^{th}$ instance. 

\begin{table}[h]
    \centering
    \begin{tabular}{|c||c|c||c|c|c|c|c||}
        \hline
        & \multicolumn{2}{c||}{\textit{Constraint Violation}} & \multicolumn{5}{c||}{\textit{Objective Statistics}}  \\
        \hline
        $\alpha$ & \% Violated & Mean Violation & \% Under  & \% Over  & Avg-O & Avg-U & AR\\ 
        \hline
        0-0.2 & 64.7 & 105.7 & 64.1 & 35.9 & 39.1 & 13.6 & 1.254 \\
        \hline
        0.2-0.4 & 66.1 & 8.77 & 56.8 & 43.2 & 3.26 & 4.49 & 1.0417 \\
        \hline
        0.4-0.6 & 95.42 & 9.9 & 14.92 & 85.1 & 3.17 & 1.36 & 1.029  \\ 
        \hline
        0.6-0.8 & 99.84 & 10.4 & 8.6 & 91.4 & 1.97 & 0.55 & 1.0185  \\
        \hline 
        0.8-1 & 100 & 7.55 & 3.17 & 96.83 & 0.66 & 0.11 & 1.0064 \\ 
        \hline
        All & 84.93 & 21.14 & 30 & 70 & 5.89 & 7.76 & 1.0703 \\
        \hline
    \end{tabular}
    \caption{Baseline Neural Network Performance}
    \label{tab KP: Baseline NN}
\end{table}

The FC model shows strong performance on the AR. Performance is relatively poor on low capacity instances. It would be expected that low capacity instances are inherently harder since the inclusion or exclusion of a single item could be the difference between a near optimal solution and a poor solution. This is reflected in the outsized mean violation on low capacity instances. This gives a distorted view of performance on these instances that is not present when the constraint violation is considered in absolute terms. Rates of constraint violation are extremely high across all quintiles, reaching 100\% in the highest capacity quintile. This reflects that the constraints are not explicitly modelled during learning. Table \ref{tab KP: Coldstart NN} reports results on the LDF model.

\begin{table}[h]
    \centering
    \begin{tabular}{|c||c|c||c|c|c|c|c||}
        \hline
        & \multicolumn{2}{c||}{\textit{Constraint Violation}} & \multicolumn{5}{c||}{\textit{Objective Statistics}}  \\
        \hline
        $\alpha$ & \% Violated & Mean Violation & \% Under  & \% Over  & Avg-O & Avg-U & AR\\ 
        \hline
        0-0.2 & 10.48 & 195 & 96.8 & 3.16 & 113.2 & 30.35 & 1.5643 \\
        \hline
        0.2-0.4 & 0.16 & 2.93 & 100 & 0 & N/A & 12.73 & 1.1521 \\
        \hline
        0.4-0.6 & 0 & N/A & 100 & 0 & N/A & 8.46 & 1.0931 \\
        \hline
        0.6-0.8 & 0 & N/A & 100 & 0 & N/A & 5.88 & 1.0628 \\
        \hline
        0.8-1 & 0 & N/A & 100 & 0 & N/A & 2.62 & 1.027 \\
        \hline
        All & 2.13 & 191 & 99.4 & 0.6 & 113.24 & 12 & 1.1811 \\
        \hline
    \end{tabular}
    \caption{LDF Neural Network Performance}
    \label{tab KP: Coldstart NN}
\end{table}

Constraint satisfaction is significantly improved in the LDF model. Across all instances, the capacity constraint is violated only 2.13\% of the time, a significant improvement on the 85\% violation rate reported on the FC model. These constraint violations exclusively occur in lower capacity instances, moderate to high capacity instance recorded no constraint violations whatsoever. This comes at some cost to the approximation ratio, but the increase is relatively minor, on the order of 10\%. Table \ref{tab KP: Hotstart NN} reports results on the pre-trained LDF model.

\begin{table}[h]
    \centering
    \begin{tabular}{|c||c|c||c|c|c|c|c||}
        \hline
        & \multicolumn{2}{c||}{\textit{Constraint Violation}} & \multicolumn{5}{c||}{\textit{Objective Statistics}}  \\
        \hline
        $\alpha$ & \% Violated & Mean Violation & \% Under  & \% Over  & Avg-O & Avg-U & AR\\ 
        \hline
        0-0.2 & 14 & 278 & 93.8 & 6.2 & 92.1 & 30.4 & 1.51 \\
        \hline
        0.2-0.4 & 0.32 & 1.93 & 100 & 0 & N/A & 2.38 & 1.1894 \\
        \hline
        0.4-0.6 & 0.17 & 1.38 & 100 & 0 & N/A & 8.78 & 1.09732 \\
        \hline
        0.6-0.8 & 0 & N/A & 100 & 0 & N/A & 5.29 & 1.0562 \\
        \hline 
        0.8-1 & 0 & N/A & 100 & 0 & N/A & 2.37 & 1.0245 \\
        \hline
        All & 2.9 & 268 & 98.8 & 1.2 & 92.1 & 12.4 & 1.177 \\
        \hline
    \end{tabular}
    \caption{Pre-trained LDF Neural Network Performance}
    \label{tab KP: Hotstart NN}
\end{table}

The performance of the pre-trained LDF model is very similar to the base LDF model, with no clear indication that either model is strictly better. As noted previously we found training to be significantly easier with the pre-trained model. In particular, it was found to be more robust with respect to hyperparameter changes, and alleviated the strong dependence on gradient clipping to deal with exploding gradients.

The computational time required for training and generating predictions is also of interest, particularly in comparison to traditional solvers. The total epochs and time taken for the three models to converge during training are listed in Table \ref{tab KP: Convergence Times}. For the pre-trained LDF model these values include the pre-training time. The LDF model takes significantly longer to converge fully. Prediction times for the neural network models, averaged over the full set of 30,000 instances, took 1ms per instance. Gurobi on the same set of instances takes on average 5.7ms per instance, noting that Gurobi is used here for convenience but other methods may be significantly faster at solving KP. Prediction time could likely be further reduced by using a GPU, with the caveat that in reality it may not be useful to solve more than one instance at a time. More generally for harder IP problems (or harder KP instances) we would expect predictions generated by neural networks to be significantly faster.

\begin{table}[h]
    \centering
    \begin{tabular}{|c|c|c|c|}
        \hline
        & \; FC \; & \; Pre-trained LDF \; & \; LDF \; \\
        \hline
        Epochs & 16 & 46 & 250 \\
        \hline
        \; Time (Minutes) \; & 0.7 & 2.6 & 16 \\
        \hline
    \end{tabular}
    \caption{Training Time for Neural Network Models}
    \label{tab KP: Convergence Times}
\end{table}

These results demonstrate an ability to trade-off optimality for constraint satisfaction by using the LDF. In LDF models poor performance is mostly concentrated in low capacity instances, moderate and high capacity instances achieve strong performance in terms of the AR without violating any constraints on the test set. It's likely that the unconstrained FC model achieves strong performance in terms of the approximation ratio by consistently violating the capacity constraint by a moderate amount. Enforcing constraint satisfaction reduces or removes these violations but does not directly help the model learn the weights in the optimal solution leading to a small increase in the AR. This gives rise to a trade-off between optimality and constraint satisfaction. It's not obvious whether this relationship would generalise to other IP problems, in fact experiments applying LDF to the job shop scheduling problems report impressive improvement in both optimality and constraint satisfaction \cite{kotary22}.

\section{Conclusions}
In this paper, we investigated using neural networks to approximate solutions to the Knapsack Problem while committing to the constraints. It experimentally demonstrated that the Lagrangian Dual Framework could be used to strongly encourage satisfaction of the knapsack capacity constraint with a reasonably small reduction in the optimality of predicted solutions. It discussed issues in interpreting neural network outputs as solutions to integer programming problems and evaluating their performance, particularly in the context of model validation. It also demonstrated that using a pre-trained neural network as a base model may alleviate problems with exploding gradients found in the LDF model. While the scope of this work is limited to KP, it is hoped that some of the principles and techniques discussed will also be applicable to the approximation of solutions for other combinatorial optimisation problems using neural networks. An obvious research direction is applying LDF to more challenging combinatorial optimisation problems. In general solving combinatorial optimisation problem instances may itself be a difficult and computationally expensive task, which makes generating training sets challenging. In focusing on KP we sidestep these issues surrounding data generation, but this will be an interesting challenge in further research.

\bibliographystyle{ieeetr}
\bibliography{references}

\end{document}